# Assessing the Quality of MT Systems for Hindi to English Translation

Aditi Kalyani
Banasthali University
Rajasthan, India

Hemant Kumud
Banasthali University
Rajasthan, India

Shashi Pal Singh
AAI, CDAC
Pune, India

Ajai Kumar
AAI, CDAC
Pune, India

## ABSTRACT
Evaluation plays a vital role in checking the quality of MT output. It is done either manually or automatically. Manual evaluation is very time consuming and subjective, hence use of automatic metrics is done most of the times. This paper evaluates the translation quality of different MT Engines for Hindi-English (Hindi data is provided as input and English is obtained as output) using various automatic metrics like BLEU, METEOR etc. Further the comparison automatic evaluation results with Human ranking have also been given.

## 1. INTRODUCTION
The main objective of MT is to break the language barrier in a multilingual nation like India. Evaluation of Machine Translation (MT) like MT itself has proven to be a very difficult task since the instigation. The difficulty arises primarily from the reason that most sentences can be translated in many acceptable ways. A lot of researchers have been discussing this issue since the beginning. Furthermore, as it was stated by Y. Wilks, ―machine translation evaluation is better understood than machine translation‖ (Carbonell, Wilks 1991) and ―machine translation evaluation is a better founded subject than machine translation‖ (Wilks 1994).

The situations worsen especially when we are talking about Indian languages, which have relatively free order, the translation as well as assessment becomes more difficult. Consequently, there is no specific standard against which a translation can be evaluated for appropriateness because Human assessments are highly subjective and automatic metrics are unable to cover all the aspects of translation.

MT Evaluation strategies were initially proposed by Miller and Beeber-center in 1956 followed by Pfaffine in 1965. In the beginning MT evaluation was carried out only by human judges. This process, however, was time-consuming and highly subjective. Then as the field of machine Translation grew there arose the dire need for automation i.e., for fast, objective, and reusable methods of evaluation, the results of which are not biased or subjective at all. To this date, several metrics for automatic evaluation have been proposed and which are accepted by the MT community enthusiastically.

This paper discusses and compares the implementation results of various automatic metrics for Hindi-English. Section 2 provides a brief history study of human and automatic evaluation strategies. Section 3 describes automatic metrics used by us in detail. Section 4 presents the scores obtained using various metrics. Section 5 compares automatic evaluation results with Human score. Section 6 discusses the issues in handling Indian Languages. Finally Section 7 concludes the work done along with future trends.

## 2. RELATED WORK
Manual evaluation is done by calculating fluency, adequacy and fidelity (Hovy, 1999; White and O'Connell, 1994). Fluency and adequacy are measured separately on each of the translations obtained and are generally judged on a five or seven point scale (Przybocki 2008). They are occasionally averaged to give a single numerical score to the output.

Post-editing is also one of the methods of measuring translation quality where the system output is corrected after being produced, since more accurate translation require less editing. This method requires a large amount of work to be done by human annotators to correct system output, rather than just scoring it on a given scale.

By the mid 1990's, the results of regular ARPA evaluations led to uncertainties concerning the legitimacy and dependability of human ratings due to various factors. Moreover, humans required a lot of time, money and are highly subjective. Hence the researchers started exploiting the idea of fully automatic measures seriously in 1990.

The boom of automatic metric started with the introduction of BLEU (Papineni et al., 2001) which is based on average of matching n-grams between candidate and reference. This was as if a panacea to all the problems in evaluation has been found. BLEU till this date also is considered as the default standard for automatic evaluation. Following IBM's lead NIST (Doddington, 2002) came out, which calculates matched n-grams of sentences and attach different weights to them. GTM (Turian et al., 2003) computes precision, recall and f-measure in terms of maximum unigram matches. In same year ROUGE (Lin and Hovy, 2003) was introduced that created the summary & compared it with the summary created by human (Recall oriented).2005 proved to be very important because one of the most successful metric METEOR (Banerjee & Lavie, 2005) {latest modification: 2012} released. This was based on various modules (Exact Match, Stem Match, Synonym Match and POS Tagger). After this various other metrics came into existence and many versions of already existing metrics were also released.

Some of the other automatic metrics are:

- BLANC (Lita et al., 2005): Based on features of BLEU and ROUGE

- TER (Snover et al., 2006): Metric for measuring mismatches

- ROSE (Song and Cohn, 2011): Uses syntactic resemblance (Here Part of Speech)

- AMBER (Chen and Kuhn, 2011): Based on BLEU but adds recall, extra penalties , and some text processing variants

- LEPOR (Han et al., 2012): Combines sentence length penalty and n-gram position difference penalty. Also uses precision and recall

- PORT (Chen et al., 2012): Based on precision, recall, strict brevity penalty, strict redundancy penalty and an ordering measure.

- METEOR Hindi (Ankush Gupta et al., 2010): A modified version of the METEOR containing features specific to Hindi

In spite of existence of so many metrics there is no such metric which works such that it can correlate well with humans and can be used on all the languages (esp. free word order languages).

## 3. AUTOMATIC EVALUATION

Mostly all automatic metrics are based one of the following methods to calculate scores [14]:

- **Edit Distance Based**: Number of changes required to make candidate as reference in terms of number of insertions, deletions and substitutions are counted

- **Precision Based:** Total number of matched unigrams are divided by the total length of candidate

- **Recall Based:** Total number of matched unigrams is divided by the total length of reference

- **F-measure Based:** Both precision and recall scores are used collectively

Below are discussed a few metrics which are used by us for evaluation.

### 3.1 BLEU

In 2000, Papineni proposed BLEU (Bilingual Evaluation Understudy) metric in IBM. BLEU is one of the n-gram based metrics. Here for each n (where n usually ranges from 1 to a maximum of 4), number of n-grams in the test translation that have a match in the corresponding reference translations are counted.

BLEU uses modified n-gram precision in which a reference translation is considered exhausted after a matching candidate word is found. A brevity penalty is introduced to compensate difference in the length of candidate and reference translations.

The final BLEU formula [1] is:

$$S_{\text{bleu}} = e^{(1-\frac{r}{t})} \; e^{\sum_{i=1}^{N} w_i \log p(s_i)}$$

Since the precision of 4-gram is many times 0, the BLEU score is generally computed over the test corpus rather than on the sentence level. Many enhancements have been done on the basic BLEU algorithm, e.g. Smoothed BLEU (Lin and Och 2004) etc. to provide better results.

### 3.2 NIST

Doddington established NIST in 2002 which is similar to BLEU except, it assigns a weight to each unigram depending upon its uniqueness or how informative the n-gram is. It uses Arithmetic mean rather than geometric mean. The formula for NIST can be given as, [3]

$$\text{Score} = \sum_{i=1}^{n} \left\{ \sum_{\substack{all \; w_1 \dots w_n \\ that \; co-occur}} \frac{p(w_1 \dots w_n)}{\sum_{\substack{all \; w_1 \dots w_n \; in \\ system \; output}}} \right\} \cdot$$
$$\exp\left\{ \beta \; log^2 \left[ min\left(\frac{L_{sys}}{L_{ref}}, 1\right) \right] \right\}$$

The information weight is given as,

$$\text{Info} \, (w_1 \dots w_n) = log_2 \left( \frac{the \; \# \; of \; occurences \; of \; w_1 \dots w_{n-1}}{the \; \# \; of \; occurences \; of \; w_1 \dots w_n} \right)$$

### 3.3 GTM

Turian et al established GTM (General Text Matcher) that was based on the idea of Melamed et al, 2003. By sharing of matched words between MT output and reference output, the evaluation score is obtained. It is not only based on precision and recall but also on harmonic mean of precision and recall, known as F-measure [14].

$$\text{F-measure} = \frac{2PR}{P+R}$$

### 3.4 METEOR

METEOR (Metric for Evaluation of Translation with Explicit Ordering) is based on a word-to-word alignment between the machine-generated translation and the reference translation. Every unigram in the test translation should map to zero or one unigram in the reference sentence. If there are two alignments with the same number of mapping, the alignment is chosen with less number of intersections of the two mappings. The score is equal to the harmonic mean of unigram precision and unigram recall.

Original METEOR consists of: [4]

1) Exact Match mapping words that are exactly same;

2) Stem Match links words that share the same stem;

3) Synonym Match mapping unigrams that are synonyms of each other.

METEOR-Hindi includes following additional modules to make more efficient for Hindi: [2]

1) The local word group (LWG) consisting of a content word and its associated function words;

2) Clause Match- Clause is defined as a phrase containing at least a verb and a subject;

3) POS matcher computes the number of matching words with same POS tag.

### 3.5 TER

TER (Translate Error Rate) was proposed by Snover and Dorr 2006, is a more. It represents the number of edits necessarily required to transform the machine output to reference translation, normalized on the length of the references [14].

$$\text{TER} = \frac{number \; \_of \; \_edits}{average \; \_of \; \_reference \; \_words}$$

Edits typically include insertion, deletion, substitutions of single words and shifts of word sequence (chunks).

## 4. EVALUATION RESULTS

For evaluation 10,000 sentences from varied domains have been used. These are Hindi sentences with their translations in English (Candidate) by different MT Engines given in Table 1 along with their translations provided by a Human (Reference). These sentences are divided in three documents.

**Table 1. MT Systems**

| Engine No. | Engine Name |
|------------|-------------|
| E1 | Google MT Engine[1] |
| E2 | Babylon MT Engine[2] |
| E3 | Microsoft Bing MT Engine[3] |

### 4.1 Evaluating Hindi - English Translation

Table 2 provides results for BLEU, NIST, METEOR, GTM and TER.

**Table 2. Results**

| | Doc No. | E1 | | E2 | | E3 | |
|---|---|---|---|---|---|---|---|
| | | Ref 1 | Ref 2 | Ref 1 | Ref 2 | Ref 1 | Ref 2 |
| **BLEU** | 1 | 0.28 | 0.30 | 0.39 | 0.40 | 0.27 | 0.31 |
| | 2 | 0.31 | 0.34 | 0.43 | 0.44 | 0.33 | 0.34 |
| | 3 | 0.41 | 0.42 | 0.39 | 0.39 | 0.36 | 0.38 |
| **N IST** | 1 | 0.30 | 0.32 | 0.37 | 0.39 | 0.22 | 0.26 |
| | 2 | 0.29 | 0.29 | 0.37 | 0.42 | 0.28 | 0.29 |
| | 3 | 0.36 | 0.39 | 0.35 | 0.38 | 0.32 | 0.39 |
| **METEOR** | 1 | 0.66 | 0.65 | 0.62 | 0.61 | 0.54 | 0.51 |
| | 2 | 0.69 | 0.73 | 0.64 | 0.59 | 0.45 | 0.55 |
| | 3 | 0.56 | 0.67 | 0.70 | 0.67 | 0.63 | 0.66 |
| **GTM** | 1 | 0.61 | 0.59 | 0.66 | 0.62 | 0.49 | 0.55 |
| | 2 | 0.63 | 0.67 | 0.62 | 0.60 | 0.49 | 0.52 |
| | 3 | 0.59 | 0.56 | 0.51 | 0.45 | 0.54 | 0.65 |
| **TER** | 1 | 0.39 | 0.42 | 0.45 | 0.41 | 0.36 | 0.30 |
| | 2 | 0.39 | 0.37 | 0.56 | 0.64 | 0.31 | 0.39 |
| | 3 | 0.49 | 0.48 | 0.49 | 0.43 | 0.51 | 0.59 |

---

[1] http://translate.goolge.com

[2] http://translation.babylon.com

[3] http://www.microsofttranslator.com

The outputs of MT Engines have been evaluated using 1 as well as 2 number of references.

Amongst all the above metrics, the performance of METEOR is the best followed by GTM. The above scores are based on document level for MT Engines mentioned earlier.

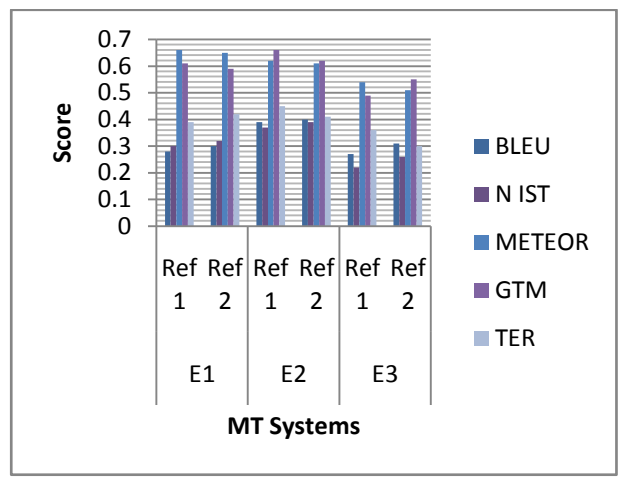

**Fig 1: Sentence Level Scores**

Figure 1 shows the comparison of various metrics for single and two references. It can be noted that in most of the cases the results for two references is better than one reference. It can be inferred that with more number of references higher accuracy and correlation can be achieved. Same can be said about the quality of reference corpus also because all the metrics compare the system output with the human reference in order to evaluate.

## 5. MANUAL Vs AUTOMATIC

Human evaluators perform evaluation based on reference translations given by human subjects. Human evaluation given here is based on 5 scales and 10 parameters [13].

### 5.1 Scale

| Rating | Translation-Quality |
|--------|---------------------|
| 1 | Unacceptable |
| 2 | Barely Understandable |
| 3 | Understandable |
| 4 | Good |
| 5 | Excellent |

### 5.2 Parameter

1. Translation of Gender and Number of the Noun/s.

2. Translation of tense in the source sentence.

3. Translation of Voice in the source sentence.

4. Identification of the Proper Nouns.

5. Use of Adjectives and Adverbs corresponding to the nouns and verbs in the source sentence.

6. Selection of proper words / synonyms.

7. The sequence of Noun, Helping Verb and Verb in the translation.

8. Use of Punctuation signs in the translation.

9. Maintaining the stress on the significant part in the source sentence in the translation.

10. Maintaining the semantics of the source sentence in the translation.

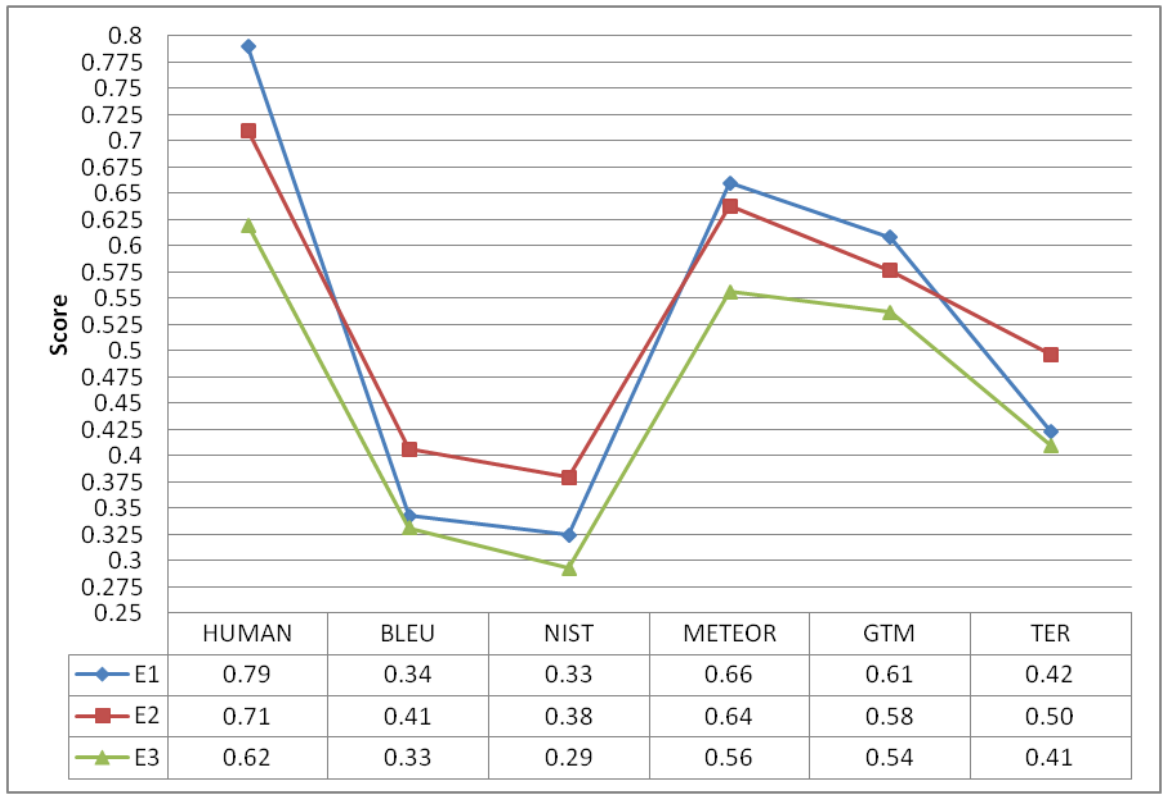

| | HUMAN | BLEU | NIST | METEOR | GTM | TER |
|---|---|---|---|---|---|---|
| E1 | 0.79 | 0.34 | 0.33 | 0.66 | 0.61 | 0.42 |
| E2 | 0.71 | 0.41 | 0.38 | 0.64 | 0.58 | 0.50 |
| E3 | 0.62 | 0.33 | 0.29 | 0.56 | 0.54 | 0.41 |

**Fig2: Comparison with Human Evaluation**

Figure 2 provides a high level view of the overall system performance i.e. how much score does a particular metric gives for a particular engine or rather which system performs better.

Here, the scores of all metrics with human evaluated scores are compared. The performance of E2 is best according to most of the metrics but as per METEOR and GTM E1's performs better than E2. But according to human the performance of E1 is better.

## 6. EVALUATING INDIAN LANGUAGES

Hindi is one of the languages which are very rich morphologically. With this the handling while translation as well as evaluation turns out to be more difficult. There are various problems when handling Hindi-English or English-Hindi translations. Some of them are:

- Difference in Word Order: Hindi is a language which follows a relatively free word order.

For E.g.: राम आम खाता है (SOV)

English (Incorrect): Ram mango eats

English (Correct): Ram eats mango (SVO)

Because of difference in word order the translation becomes difficult, as it is wrong to just translate words. The sentence is not grammatically correct and also it is semantically wrong. Also there are many correct translations of a single sentence which makes it harder to evaluate.

- Categorical divergence: When the lexical category of a word has to be changed during translation it gets very difficult to handle [7].

For E.g.: मुझे **भूख लग रही** है

English: I am **feeling hungry**

Here the noun-verb combination ("भूख लगना") changes to verb-adjective combination *feeling hungry*.

- Non-literal translation: This mostly occurs due to difference in regions, as India has diverse culture and huge number of languages.

For E.g.: मुझे खाना खाना है और पानी पीना है

मुझे भोजन करना है एवं जल पीना है

Here both the sentences mean the same but only differ in word order choices. This causes a lot of problems in evaluating translation as many metrics work only on n-gram matching. Hence require more sophisticated metrics for evaluation.

- Sense Differences: In many cases a single word translates to various words (one word with different senses) and has varied meanings.

For E.g.: मैंने एक लड़की को बाज़ार में **देखा**

मैंने एक लड़की को बाज़ार में **आरी से काटा**

English: I **saw** a girl in the market

Here the meaning of sentence completely changes and this leads to many misapprehensions.

- Pleonastic Divergence: Here the 'It' has no semantic content (such a constituent is called a pleonastic).

  For E.g.: बर्फ पड़ रही है

  English: It is snowing

  In Hindi there is no equivalent translation for the word 'it'. There are many such words that do not have corresponding translations or rather which get embedded with other words to form sentence.

  Hence handling evaluating such translations is difficult because no proper alignment is produced.

# 7. CONCLUSIONS

This paper demonstrates various approaches of evaluation. Here we provided the implementation results of various automatic evaluation metrics and compared there performances for different translation systems. Then the result of automatic metrics have been contrasted with Human evaluated results and the correlation results i.e. which metric correlates best with Human evaluated results have also been presented.

From the correlation it can be concluded that METEOR highly correlates with human judgment in most of the cases but fails in certain, this might be because of METEOR's working on superficial level. Hence deeper evaluation strategies are required. Amongst all the translation engines, E2 performs best according to most of the metrics but as per METEOR and GTM, E1 performs the best. This paper also discussed various issues with translating and evaluating Indian languages.

Even though so many automatic metrics exist, there is no single metric which can perform exceptionally well on all the language pairs and even if it does, it requires a huge corpus and other language resources which are not available as of now. Hence a metric such as this needs to be devised which can handle all the problems involved in evaluation of MT Output.